\documentclass[
]{ceurart}

\usepackage{listings}
\usepackage{graphicx}

\usepackage{booktabs, tabularx, multirow}
\usepackage{xurl}
\usepackage{float}
\graphicspath{ {./images/} }
\begin{document}

\copyrightyear{2022}
\copyrightclause{Copyright 2022 for this paper by its authors. Use permitted under Creative Commons License Attribution 4.0 International (CC BY 4.0).}

\conference{IberLEF 2022, September 2022, A Coruña, Spain.}

\title{Deep Learning Brasil at ABSAPT 2022: Portuguese Transformer Ensemble Approaches}


\author[1]{Juliana Resplande Sant'Anna Gomes}[%
orcid=0000-0001-6900-1931,
email=juliana.resplande@discente.ufg.br,
]
\cormark[1]
\fnmark[1]

\author[1]{Eduardo Augusto Santos Garcia}[%
email=edusantosgarcia@discente.ufg.br,
orcid=0000-0002-0549-8771,
]
\fnmark[1]

\author[1]{Adalberto Ferreira Barbosa Junior}[%
email=adalbertojunior@discente.ufg.br,
]

\author[2]{Ruan Chaves Rodrigues}[%
email=ruanchaves93@gmail.com,
]

\author[1]{Diogo Fernandes Costa Silva}[%
email=diogo_fernandes@discente.ufg.br,
]

\author[1]{Dyonnatan Ferreira Maia}[%
email=dyonnatan@discente.ufg.br,
]

\author[1]{Nádia Félix Felipe da Silva}[%
email=nadia@inf.ufg.br,
]

\author[1]{Arlindo Rodrigues Galvão Filho}[%
email=arlindo@inf.ufg.br,
]

\author[1]{Anderson da Silva Soares}[%
email=anderson@inf.ufg.br,
]

\address[1]{Institute of Informatics, Federal University of Goiás, Brazil}
\address[2]{Faculty of Informatics, University of the Basque Country (UPV/EHU), Spain}

\cortext[1]{Corresponding author.}
\fntext[1]{These authors contributed equally.}

\begin{abstract}
Aspect-based Sentiment Analysis (ABSA) is a task whose objective is to classify the individual sentiment polarity of all entities, called aspects, in a sentence. The task is composed of two subtasks: Aspect Term Extraction (ATE), identify all aspect terms in a sentence; and Sentiment Orientation Extraction (SOE), given a sentence and its aspect terms, the task is to determine the sentiment polarity of each aspect term (positive, negative or neutral). This article presents we present our participation in  Aspect-Based Sentiment Analysis in Portuguese (ABSAPT) 2022 at IberLEF 2022. We submitted the best performing systems, achieving new state-of-the-art results on both subtasks.
\end{abstract}


\maketitle

\section{Introduction}

Aspect-based Sentiment Analysis (ABSA) can be defined as a fine-grained approach to sentiment analysis. In this task, instead of attempting to classify an entire sentence under a single sentiment polarity, we must classify the individual sentiment polarity of all tokens that make a significant contribution to the overall sentiment of the sentence. In the task terminology, these tokens are called the aspect terms of a sentence. 

Inspired on the format of SemEval-2014 Task 4 \cite{pontiki-etal-2014-semeval}, the Aspect-Based Sentiment Analysis in Portuguese (ABSAPT) 2022 at IberLEF 2022 features two subtasks. The first subtask is called Aspect Term Extraction (ATE): given a set of sentences, the task is to identify all aspect terms present in each sentence. In this article, we present our participation at the Aspect-Based Sentiment Analysis in Portuguese (ABSAPT) 2022 at IberLEF 2022. We submitted the best performing systems, achieving new state-of-the-art results on all subtask.

The second subtask is called Sentiment Orientation Extraction (SOE): given a set of sentences that have already been annotated for their aspect terms, the task is to determine the sentiment polarity of each aspect term (positive, negative or neutral). This subtask is also known as Aspect Term Polarity Classification (ATP) \cite{pontiki-etal-2014-semeval, toh2014dlirec} or Aspect Sentiment Analysis (ASA) \cite{8976252}.


The remainder of this article is structured as follows: In the next section, \textbf{Related Work}, we go into the previous research that supported our approach; Next, under \textbf{Dataset}, we present a detailed analysis of the training data provided for both subtasks. Moreover, under \textbf{Methodology}, \textbf{Experimental setup} and \textbf{Results}, we articulate our strategy to address each subtask; Finally, under \textbf{Conclusion}, we bring an overview of the results and a proposal for future work.


\section{Related Work}

ATE ABSITA \cite{de2020ate} was the EVALITA 2020 \cite{basile2020evalita} shared task on Aspect Term Extraction and Aspect-Based Sentiment Analysis. Both the first-ranked team \cite{di2020app2check} and the second-ranked team \cite{bennici2020ghostwriter19} had the approach of simply framing Aspect Term Extraction as a Named Entity Recognition task, and then fine-tuning state-of-the-art Transformer models on the training data for the task. We followed a similar approach during our participation at ABSAPT 2022.   

ATE ABSITA \cite{de2020ate} also features a subtask similar to Sentiment Orientation Extraction (SOE) at ABSAPT 2022. The first-ranked team \cite{di2020app2check} framed it as a problem of text classification, under the premise that the portion of the text that surrounds each aspect should have the same overall sentiment as the aspect itself.

Although we also experimented with this approach, our best performing system at the Sentiment Orientation Extraction subtask framed it as a text generation problem, similar to what was done by Zhang et al. \cite{zhang2021towards} and Chebolu et al. \cite{chebolu2021exploring}.


\section{Dataset}

The dataset was taken from TripAdvisor reviews, specifically from the hospitality industry, consisting of hotel and room reviews.	Table \ref{tab:SOE} is an example of a positive polarity from the train dataset.  The structure of the data is as follows: id, review, polarity, aspect, start position and end position.

\begin{table}[ht]
\caption{Sentiment Orientation Extraction example. Start and End positions are abbreviated into start pos. and end pos.}
\centering
\begin{tabularx}{\textwidth}{lXcccc}
\toprule
id   & review                                                                                                         & polarity & aspect   & start pos. & end pos. \\ \midrule
2414  & Hospedei-me em maio nesse hotel  \newline pela terceira vez ...                                                                & 1        & hotel   & 26            & 31          \\ \bottomrule

\end{tabularx}

\label{tab:SOE}
\end{table}

For the Aspect Term Extraction task, we build a NER dataset converting the original train data to the BIO/IOB format (Inside, Outside, Beginning), a common tagging format where each token can be classified with the prefixes \emph{B}, \emph{I} and \emph{O}. The \emph{B} prefix indicates the begging of a new classification chunk, the \emph{I} prefix indicates that the token is inside a previous chunk and the \emph{O} tag indicates that the token doesn't belong to any class or chunk. An example of the annotation is shown in Table \ref{tab:ATE}.
\begin{table}[ht]
\caption{Example of an Aspect Term Extraction sentence converted to the BIO tagging format.}
\centering

\resizebox{\columnwidth}{!}{%
\begin{tabular}{|c|c|c|c|c|c|c|c|c|c|c|c|c|c|c|}
\hline
A & estrutura & do & hotel    & é & muito & boa. & A & piscina  & é & excelente & e & os & quartos  & também. \\ \hline
O & O         & O  & B-ASPECT & O & O     & O    & O & B-ASPECT & O & O         & O & O  & B-ASPECT & O      \\
\hline
\end{tabular}
}
\label{tab:ATE}
\end{table}

\begin{figure}[ht]
    \centering
    \includegraphics[width=0.7\textwidth]{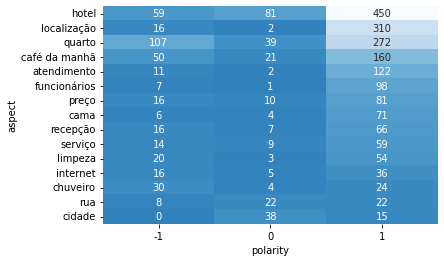}
    \caption{Aspect polarity heat map from top 15 ocorrences.}
    \label{fig:polarityHeadMap}
\end{figure}

Analyzing the provided dataset in search of imbalanced data that could be exploited, we found some noteworthy cases.
Polarity, for example, is unbalanced towards 1 (positive), representing about 68\% of the dataset. Furthermore, the distributed polarity with respect to the aspects is also unbalanced, as seen in Figure \ref{fig:polarityHeadMap}.

\begin{figure}[ht]
    \centering
    \includegraphics[width=0.75\textwidth]{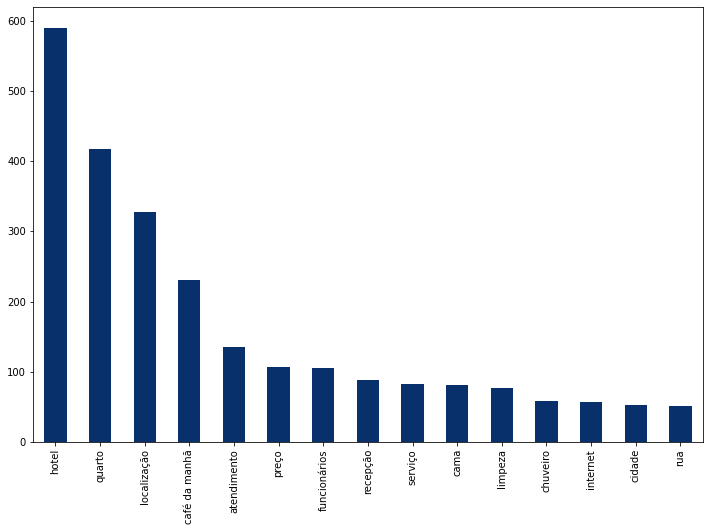}
    \caption{The number of occurrences of the 15 most common aspect occurrences in the training dataset.}
    \label{fig:aspectOccurrences}
\end{figure}

Another interesting point is the aspect term occurrences. Although the dataset contains 77 unique aspects, the top 15 represent 79\% of the data, as shown in Figure \ref{fig:aspectOccurrences}. Furthermore, the aspect term position is not evenly distributed across the dataset, as its occurrence is more common at the beginning of a review.

\begin{figure}[ht]
    \centering
    \includegraphics[width=0.75\textwidth]{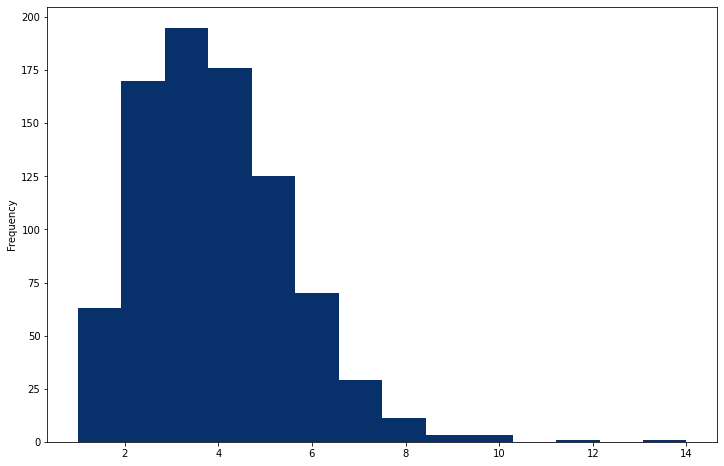}
    \caption{The frequency distribution of the total number of aspects for each unique sentence on the training dataset.}
    \label{fig:aspectFrequency}
\end{figure}

Concerning the Aspect Term Extraction, we found that the distribution of aspects in the reviews has a mean of 3.7 and a standard deviation of 1.7, As seen in Figure \ref{fig:aspectFrequency}. In contrast, the distribution of words has a mean of 68 and a standard distribution of 21. The low amount of aspects per review can provide an extra challenge for a model to learn accurately, since most of the words in a given review will not be an aspect.

\section{Methodology}
We adopt different methodologies in the ATE and SOE tasks. For ATE, we treat it primarily as a single sentence tagging task \cite{https://doi.org/10.48550/arxiv.1810.04805}. For SOE, we test two distinct strategies: treating SOE as sentence pair classification or as conditional text generation. 

\subsection{Task 1 - Aspect Term Extraction}
\label{subsec:metolodogy-ATE}

For Aspect Term Extraction, we assess four training considerations: training strategy, Transformer model, dataset preprocessing and dataset configuration, in which we explain in the next subsections. 

\subsubsection{Training strategy}
For the objective of Aspect Term Extraction, we transform the dataset to behave as a Named Entity Recognition task (NER) \cite{di2020app2check}, where we classify chunks of the sentence with the objective to identify which group of tokens represents an aspect. The reviews are tokenized and each aspect is tagged with the BIO format, we train a transformer model using the Hugging Face transformers library \cite{https://doi.org/10.48550/arXiv.1910.03771}.

\subsubsection{Transformer model}
\label{subsubsec:models-config-ATE}
We consider evaluating the following Tranformer-based models with Portuguese support: Bertimbau base ; mDeBERTa v3 base; and our own RoBERTa base in Portuguese trained on BrWaC and the Portuguese portion of OSCAR corpora.
Also an mDeBERTA pretrained model on Evalita, MAMs and Semeval datasets, which can be considered as mDEBERTA trained previously on additional data.

\subsubsection{Dataset preprocessing} 
The original competition dataset is separated by aspect, for the conversion to a NER dataset we join all the aspects of reviews with the same text and each unique review it's used as a single training example. We use the NTLK library \cite{loper2002nltk} to tokenize and generate the BIO annotated dataset for a model's training.

\subsubsection{Dataset configuration}
\label{subsubsec:dataset-config-ATE}
In addition to the dataset in Portuguese of the ABSAPT 2022 we used external ABSA datasets such as Evalita, MAMs, Semeval 2014, 2015, and 2016 tasks. 

\subsection{Task 2 - Sentiment Orientation Extraction}
\label{subsec:metolodogy-SOE}
For Sentiment Orientation Extraction, we assess four training considerations: training strategy, review preprocessing, Transformer model and dataset configuration, in which we explain in the next subsections. 


\subsubsection{Training strategy}
SOE can be targeted with masked language models or autoregressive language models \cite{zhang2021towards, chebolu2021exploring}. 

If SOE is targeted with a masked language model, following Devin et. al. \cite{https://doi.org/10.48550/arxiv.1810.04805}, the review and the aspect term are concatenated with a sentence separator token ( [SEP] ), and then the resulting sequence of tokens is assigned a sentiment polarity. 

However, if SOE is targeted with an autoregressive language model, it will generate sentiment polarity labels starting from a prompt in the format of ``Review: [review content] Aspect: [aspect term] Polarity:'', which should be completed with either ``positive'', ``negative'' or ``neutral''. This is done either in a zero-shot fashion, in the case of GPT-3, or after fine-tuning a sequence-to-sequence Transformer model to examples given in this format, such as PTT5 \cite{https://doi.org/10.48550/arxiv.2008.09144}.

\subsubsection{Review preprocessing} We tested two review preprocessing setups for the SOE task. In the first one, the entire review is concatenated with the aspect term, just as described under our training strategy. In the second one, only the portion of the review that is relevant to the aspect term is concatenated with it. Following Rosa et al. \cite{di2020app2check}, we implement this by concatenating with the aspect term only the sentence in which the aspect term is located.

\subsubsection{Transformer model}
In order to determine the Transformer model to be used, we consider Tranformer models with Portuguese support: Bertimbau base; Bertimbau large; PPT5 base; PTT5 large; mDeBERTa base; XLM-RoBERTa base; LaBSE, Canine-c, RemBERT.

\subsubsection{Dataset configuration}
\label{subsubsec:dataset-config}

We test whether to use external data using B2W dataset and ABSA datasets, such as Evalita, MAMs, Semeval 2014, 2015, and 2016 tasks, by extending pseudo-training subsets on \ref{subsec:experiment-SOE}. 

We additionally tested mDeBERTA pretrained model on MNLI and XNLI provided by \texttt{MoritzLaurer/mDeBERTa-v3-base-mnli-xnli}, which can be considered as mDEBERTA trained previously on additional data: MNLI and XNLI.

Moreover, we attempt to perform data augmentation over the original task dataset throuh target swap, according to what has been suggested by Liesting et al. \cite{https://doi.org/10.48550/arxiv.2103.15912}. This approach consists of artificially increasing the amount of sentences on the dataset by replacing aspect terms by others belonging to the same category. As there are no annotated categories to the reviews on the ABSAPT 2022 dataset, we tried to approximate such categories through topic modeling. 

\section{Experimental setup}
In experiments, two V100 GPUs (32 GB) were employed, one for each task.

\subsection{Task 1 - Aspect Term Extraction}
\label{subsec:experiment-ATE}
To evaluate the quality of the models we created a pseudo test-training splits of 30/70 of training set where we make sure that each split has unique sentences, this avoids leakage of data between the pseudo-train and pseudo-test subsets. Including the external data, we create two subsets for the training of the ATE task:
    \begin{itemize}
    \item \textbf{Portuguese subset}: 70\% random split of the original training-set of the competition dataset. 
    \item \textbf{multilingual subset}: All external data from Evalita, MAMs, Semeval, plus the Portuguese subset.
\end{itemize}
All models are fine-tuned in using the HuggingFace Transformer library with a batch-size of 8, learning-rate of 3e-5 (BERTimbau, RoBERTa) or 4e-5 (mDeBERTa) for 8 epochs. The evaluation occurred in the 30\% random split of the original training-set of the competition dataset.

\subsection{Task 2 - Sentiment Orientation Extraction}
\label{subsec:experiment-SOE}

We evaluate the approaches mentioned under subsection \ref{subsec:metolodogy-SOE} by splitting the training set for the shared task into a new training set and validation set for evaluation purposes. We experiment with three different ways of splitting it:

\begin{itemize}
    \item \textbf{subset 1}: Random approach. The training set for the shared task is arbitrarily split into a new training and validation set. 
    \item \textbf{subset 2}: We are careful to keep the same proportion of reviews of each polarity on each split.
    \item \textbf{subset 3}: Besides polarity, we also try to keep the same ratio of aspect terms between the splits.
\end{itemize}

On all subsets, we attribute 70\% of the reviews to the new training set, and 30\% of the reviews to the validation set.


As shown in section \ref{sec:results}, the best results for the SOE task in our experiments were achieved by PTT5 Large with the conditional text generation training strategy, while taking the entire review and the aspect term as an input to the model, and without using any external data.

We take this model and fine-tune it under four distinct combinations of learning rate and random seed: $\{3e-4, 7\}$, $\{1e-4, 5\}$ and $\{5e-5, 8\}$. Afterward, we produce the final submission through a majority voting ensemble of the predictions of the four fine-tuned models.

Ensemble training code is available on \url{https://github.com/ju-resplande/dlb_absapt2022}.

\section{Results}
\label{sec:results}
\subsection{Task 1 - Aspect Term Extraction}

We evaluate the training strategies in \ref{subsec:metolodogy-ATE}, in terms of the following metrics: accuracy (acc.), precision-macro (precision), recall-macro (recall) and  f1-macro (f1) for the pseudo-test split of the competition dataset.

The results of the internal evaluation in the pseudo-test splits of each combination of model and dataset tested can be found in Table \ref{tab:results-ATE}.

\begin{table}[ht]
    \caption{Best results on the ATE task. The symbol \{MAMs, Evalita, Semeval\}$^*$  refers to mDeBERTa previously trained on external datasets, explained in \ref{subsubsec:models-config-ATE}.}
    \centering
    \begin{tabular}{lccccc} 
        \toprule
        model                           & external data                              & acc.                     & precision                & recall                            & f1                                 \\ 
        \midrule
        BERTimbau base                  & -                                          & 98.2                     & 78.1                     & 87.8                              & 82.6                               \\
        RoBERTa PT base                 & -                                          & 98.4                     & 80.8                     & 90.7                              & 85.5                               \\
        mDeBERTa base                   & MAMs, Evalita, Semeval                    & 98.4                      & 79.1                     &\textbf{94.0}                      & \textbf{85.9}  \\
        mDeBERTa base             & \{MAMs, Evalita, Semeval\}$^*$                     & \textbf{98.5}            & \textbf{81.4}            & 90.1                              & 85.5                               \\
    \bottomrule
    \end{tabular}
    \label{tab:results-ATE}
\end{table}

The final submission was created using an ensemble of the 3 best models, using a simple median of the label probabilities output by the models for each token in a review, the competition results for the ATE task can be found in table \ref{tab:results-competition-ATE}.

\begin{table}[htb]
    \caption{Competition final results for the Task 1 (ATE).}
    \begin{tabular}{lc} 
        \toprule
        team\_name             & acc                           \\ 
        \midrule
        \textbf{TeamDeepLearningBrasil} & \textbf{67.1448}             \\
        Teampiln               & 65.4974                      \\
        TeamUFSCAR             & 59.3715                      \\
        TeamPeAm               & 33.8243                      \\
        TeamMachadoPardo       & 22.1050                      \\
        TeamUFPR               & 17.1908                      \\
        TeamOwl                & 2.6265                       \\
        \bottomrule
    \end{tabular}
    \centering
    \label{tab:results-competition-ATE}
\end{table}

\subsection{Task 2 - Sentiment Orientation Extraction}
We evaluate the training strategies in \ref{subsec:metolodogy-SOE}, according to subsets \ref{subsec:experiment-SOE}, in terms of the following metrics: accuracy (acc.), f1-macro (f1), and f1 on each class; positive - f1(pos), neutral - f1(neu), and negative - f1(neg).

Table \ref{tab:subsets} illustrates 3 best results for each training subset described under subsection \ref{subsec:experiment-SOE}, and also has zero-shot results for GPT-3 \cite{https://doi.org/10.48550/arxiv.2005.14165}, which has not been fine-tuned on any of our data. The best results for each specific metric are highlighted in bold. 

\begin{table}[ht]
    \caption{Best results on training subsets. We also include zero-shot results obtained with GPT-3 \cite{https://doi.org/10.48550/arxiv.2005.14165} . The symbol \{MNLI, XNLI\}$^*$  refers to mDeBERTa previously trained on MNLI and XNLI, explained in \ref{subsubsec:dataset-config}.}
    \centering
    \begin{tabular}{clcccccc} \toprule
        subset & model &  external data & acc. & f1 & f1(pos) & f1(neu) & f1(neg) \\ \midrule
        \multirow{4}{*}{subset 1} & PTT5 large &   -  & \textbf{86.8}  & \textbf{78.8} & \textbf{92.6} & 62 & \textbf{81.7}  \\
        & PTT5 large  & target swap   & 86.3 & 78.3 & 92.4 & 63 & 79.6 \\
        & PTT5 base  & target swap  & 86.3 & 78.2 & 92.6 & \textbf{63.1} & 78.8 \\
        & GPT-3 & - & 80.0 & 66.0 & 90.0 & 36.0 & 73.0 \\
        \midrule
        \multirow{3}{*}{subset 2} & mDeBERTa base &   \{MNLI, XNLI\}$^*$  & \textbf{85.1} & 75.9 & \textbf{92.4} & \textbf{57.1} & 78.2 \\
        & mDeBERTa base  &  MAMs, Evalita, Semeval   & 84.6 & \textbf{76.6} & 91.6 & 55.2 & \textbf{83} \\
        & mDeBERTa base  & -  & 83.3 & 73.5 & 91.2 & 52.1 & 77.1\\
        \midrule
        \multirow{3}{*}{subset 3} & PTT5 large &   -  &  \textbf{77.4} & \textbf{75.6} & \textbf{82.3} & 60.6 & \textbf{83.8 }\\
        & mDeBERTa base  &  MAMs, Evalita, Semeval   & 76.9 & \textbf{75.6} & 82 & \textbf{62} & 82.8 \\
        & mDeBERTa base  & \{MNLI, XNLI\}$^*$  & 74.2 & 72.1 & 81.6 & 58.1 & 76.7\\
        \bottomrule
    \end{tabular}
    \label{tab:subsets}
\end{table}

\subsubsection{GPT-3}

It is important to mention that we did not translate the reviews into English before turning them into prompts for GPT-3 through the OpenAI's text completion API endpoint \footnote{\url{https://beta.openai.com/docs/api-reference/completions}}. The prompts were designed as described on subsection \ref{subsec:metolodogy-SOE}. Furthermore, due to time constraints, we did not perform predictions for the entire test set, but only for a subset of 391 reviews.

\subsubsection{Review preprocessing}

For all the best models tested, we achieved the best results by taking the entire review and the aspect term as an input, instead of trying to extract from the review only the sentences which are relevant to the aspect term.


\subsubsection{PTT5 Large}

On subset 1 and subset 3, PTT5 large without external data provides the best results, and mDeBERTa base surpasses PTT5 large only on subset 2. Therefore, in our final submission, we opted for a conditional text generation approach where we build a voting ensemble of PTT5 large models, as described under subsection \ref{subsec:experiment-SOE}.


Table \ref{tab:results-competition-SOE} shows the final results of the competition for the SOE task and the metrics of our final submission in comparison with other teams.

\begin{table}[hb]
    \caption{Competition final results for the Task 2 (SOE).}
    \centering
    \begin{tabular}{lcccc} 
        \toprule
        team\_name             & bacc                         & f1                           & precison                                           & recall                                                                                  \\ 
        \midrule
        \textbf{TeamDeepLearningBrasil} & \textbf{82.3756}            & \textbf{81.7988}            & \textbf{81.3144}                                  & \textbf{82.3756}                                                                       \\
        Teampiln               & 78.8619                     & 77.4794                     & 76.5911                                           & 78.8619                                                                                \\
        TeamUFSCAR             & 62.8992                     & 61.2248                     & 65.5697                                           & 62.8992                                                                                \\
        TeamPeAm               & 62.8992                     & 61.2248                     & 65.5697                                           & 62.8992  \\
        TeamUFPR            & 62.8992                    & 61.2248                     & 65.5697                                          & 62.8992                                                            \\
        TeamOwl                & 53.5995 &     57.2803 & 68.9396                       & 53.5996                                                            \\
        \bottomrule
    \end{tabular}
    \label{tab:results-competition-SOE}
\end{table}

\section{Conclusion}
We submitted the best performing system on ABSAPT 2022 at IberLEF 2022, achieving new state-of-the-art results on both ATE and SOE tasks. For ATE, we used an ensemble of RoBERTa and mDeBERTa's models trained in Portuguese and multilingual datasets, respectively. For SOE, we employed a voting ensemble of PTT5 large without external data.

In future work, we plan to experiment with a coreference resolution step to improve aspect term extraction by removing ambiguity from the reviews \cite{chauhan2022mixed}. It may also be interesting to consider fine-tuning Transformer models for both subtasks simultaneously through a multi-task learning framework \cite{ https://doi.org/10.48550/arxiv.1808.09238}. 
\begin{acknowledgments}
This work has been supported by the AI Center of Excellence (Centro de Excelência em Inteligência Artificial – CEIA) of the Institute of Informatics at the Federal University of Goiás (INF-UFG). 
\end{acknowledgments}

\bibliography{old-bibliography}

\end{document}